%% file: main.tex
\documentclass[runningheads]{llncs}
\usepackage[T1]{fontenc}
\usepackage{graphicx}
\usepackage{orcidlink}
\usepackage{verbatim}
\usepackage{hyperref}

\begin{document}
\title{On the effectiveness of multimodal privileged knowledge distillation in two vision transformer based diagnostic applications}

\titlerunning{Effectiveness of MMPKD in ViTs}

\author{
Simon Baur\inst{1}\orcidlink{0009-0009-4307-3078}\textsuperscript{*}, 
Alexandra Benova\inst{1,2}\orcidlink{0009-0005-0953-955X}, 
Emilio Dolgener Cantú\inst{1}\orcidlink{0009-0003-6865-0894}, 
Jackie Ma\inst{1}\orcidlink{0000-0002-2268-1690}
}

\authorrunning{Simon Baur et al.}

\institute{Fraunhofer Heinrich-Hertz-Institut, 10587 Berlin, Germany 
\and Universität Osnabrück, 49074 Osnabrück, Germany \\
    \email{simon.baur@hhi.fraunhofer.de}}

\maketitle              
\begin{abstract}
Deploying deep learning models in clinical practice often requires leveraging multiple data modalities, such as images, text, and structured data, to achieve robust and trustworthy decisions. However, not all modalities are always available at inference time. In this work, we propose multimodal privileged knowledge distillation (MMPKD), a training strategy that utilizes additional modalities available solely during training to guide a unimodal vision model. Specifically, we used a text-based teacher model for chest radiographs (MIMIC-CXR) and a tabular metadata-based teacher model for mammography (CBIS-DDSM) to distill knowledge into a vision transformer student model. We show that MMPKD can improve the resulting attention maps' zero-shot capabilities of localizing ROI in input images, while this effect does not generalize across domains, as contrarily suggested by prior research. 

\end{abstract}

\input{chapters/01_intro_and_method}

\input{chapters/02_experimental_setup_and_metrics}

\input{chapters/03_results}
\input{chapters/04_conclusion}

\input{chapters/05_funding}

\bibliographystyle{splncs04}
\bibliography{references}

\end{document}

%% file: chapters/01_intro_and_method.tex
\section{Introduction and Method}

A typical task of deep learning models in medical applications is the classification of an image $x$ (e.g. chest x-rays) to find a diagnosis $y$. In practice, a clinical professional writes a report $x^*$ post-diagnosis including additional information. Hence, $x$ and $x^*$ share a common label $y$, yet only $x$ is available when diagnostic models can effectively be deployed. Therefore $x^*$ can be considered privileged information (PI). Knowledge distillation through privileged information (KDPI)\cite{LopezPaz2016UnifyingDA,vapnik2015learning} has been shown to be effective in unimodal cases. We extend KDPI to reflect the described multimodal setting, resulting in multimodal privileged knowledge distillation (MMPKD) and evaluate it for two clinical real world tasks,  chest x‑ray \cite{johnson2019mimic} and breast cancer \cite{lee2017curated} classification. We assess vision transformer attention maps for capabilities of ROI localization, with prior research acknowledging their explanatory potential \cite{wiegreffe2019attention} as well as pitfalls of qualitative heatmap evaluation \cite{chung2024evaluating,jain2019attention}. Our contribution can be summarized as follows:

\begin{itemize}
    \item We show that MMPKD is able to significantly increase transformers' zero-shot capabilities to localize ROI in input images through attention maps.
    
    \item We show that attention maps in general remain subject to high standard deviations when evaluated for localization of key features.
    
    \item We show that the effectiveness of MMPKD in real-world medical applications is highly dependent on the dataset and other local conditions.
\end{itemize}

In the first step of MMPKD a teacher model $f_t$ is trained on $x^*$ to predict $y$. In a second step, $f_t$ is frozen and its predictions $s$ are used as soft labels to guide the student model $f_s$. Soft labels $s$ are added during \textit{training} only, while \textit{inference}  is performed solely on $x$, effectively making training multimodal while inference is unimodal. The setup can be formalized as: 

\begin{enumerate}
    \item Train teacher $f_t$ on $(x_i^*, y_i)$,
    \item Train student $f_s$ on $(x_i, y_i)$ and $(x_i, s_i)$, where $s_i$ are soft labels provided by frozen teacher $f_t$.
\end{enumerate}

The student minimizes:
\begin{equation}
\label{distill}
f_s = \arg\min \frac{1}{n} \sum_{i=1}^{n} \left[ (1 - \lambda)\, L(y_i, \hat{y}) + \lambda\, L(s_i, \hat{y}) \right]
\end{equation}

with predictions $\hat{y} = \sigma(f_s(x_i))$ and soft labels defined as:
\begin{equation}
\label{softlabels}
s_i = \sigma \left( f_t(x_i^*) \:/\: T \right)
\end{equation}
where $L$ is a loss function, $n$ is the batch size, $\sigma$ is an activation function, $\lambda \in [0, 1]$ a weighting parameter and $T$ a temperature parameter.

%% file: chapters/02_experimental_setup_and_metrics.tex
\section{Experimental Setup and Metrics}
\begin{figure}[]
    \centering
    \includegraphics[width=\linewidth]{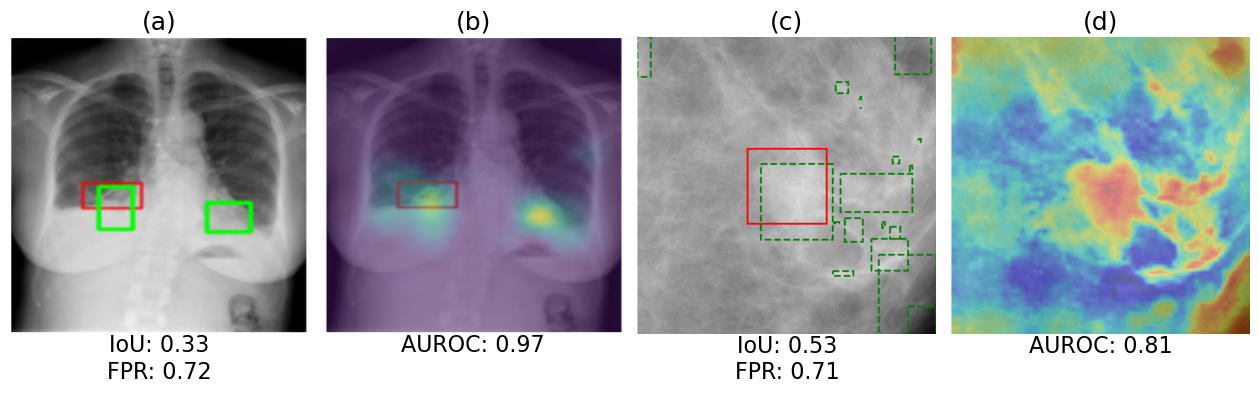}
    \caption{(a) MIMIC-CXR Sample (b) Attention map overlay (c) CBIS-DDSM mass sample (d) Attention map overlay. ROI ground truth bounding boxes are depicted red, predicted bounding boxes green. }
    \label{fig:attn_examples}
 
\end{figure}
We use official MIMIC-CXR and CBIS-DDSM datasplits. We use PubMedBERT \cite{gu2021domain} fine-tuned on MIMIC-CXR reports (excluding label names) as $f_t$ for chest X-rays (AUROC 0.99), and a random forest on CBIS-DDSM metadata as $f_t$ for mammography (AUROC 0.86/0.90 for mass/calcification) and compare to a unimodal baseline trained on images only for both datasets. All results are averaged over 5 runs with different random seeds. The student model $f_s$ for all experiments is a ViT-Tiny pretrained on ImageNet. We performed a grid search for $\lambda$ and $T$ on validation predictive performance, without observing major differences for different configurations. We evaluate: Object detection AUROC (pixel-wise between attention maps and binarized masks), IoU (best-matching box per image), and false positive rate ($\%$ of non-overlapping area of all predicted boxes with ground truth). Fig.~\ref{fig:attn_examples} illustrates the evaluation setup.

%% file: chapters/03_results.tex
\section{Results}
\begin{table}[t]
\centering
\small
\setlength{\tabcolsep}{10pt} 
\resizebox{\columnwidth}{!}{%
\begin{tabular}{l lccc}
\hline
\textbf{Dataset} & \textbf{Method} & \textbf{Obj. AUROC} & \textbf{IoU} & \textbf{FPR} \\
\hline
Mimic CXR & Baseline & $0.66 \pm 0.03$ & $0.04 \pm 0.03$ & $0.86 \pm 0.08$ \\
Mimic CXR & MMPKD    & $^*0.72 \pm 0.04$ & $^{**}0.15 \pm 0.04$ & $^{**}0.76 \pm 0.07$ \\
Mass & Baseline & $0.44 \pm 0.14$ & $0.06 \pm 0.03$ & $0.87 \pm 0.06$ \\
Mass & MMPKD    & $0.37 \pm 0.09$ & $0.03 \pm 0.03$ & $0.92 \pm 0.06$ \\
Calcification & Baseline & $0.49 \pm 0.17$ & $0.07 \pm 0.05$ & $0.73 \pm 0.13$ \\
Calcification & MMPKD    & $0.44 \pm 0.12$ & $0.04 \pm 0.03$ & $0.79 \pm 0.09$ \\
\hline
\end{tabular}%
}
\caption{Evaluation of ViT attention maps on the MIMIC, MASS, and CALC datasets. $^*$ Indicates improvement over baseline, $^{**}$ Indicates statistically significant improvement over baseline.}
\label{tab:metrics}
\end{table}

As shown in Tab.~\ref{tab:metrics}, MMPKD improves object detection performance on MIMIC-CXR over the baseline, with statistically significant gains in IoU and FPR (Mann--Whitney U test, $\alpha = 0.05$). For CBIS-DDSM (mass and calcification), no improvement or significance is observed. Predictive performance (AUROC) remains stable across all settings: MIMIC-CXR ($0.77 \pm 0.01$ for both), CBIS-DDSM mass ($0.62 \pm 0.04$ vs.~$0.64 \pm 0.01$), and calcification ($0.62 \pm 0.00$ vs.~$0.62 \pm 0.01$).

%% file: chapters/04_conclusion.tex
\section{Conclusion \& Future work}
\label{sec:conclusion}
We introduced MMPKD, a multimodal adaptation of KDPI, and evaluated it in multimodal radiology and mammography settings. While we were able to show that adding privileged information during training can significantly enhance zero-shot localization of ROI without compromising predictive performance, the benefits are task-dependent and do not consistently translate across domains, contrary to prior findings\cite{LopezPaz2016UnifyingDA}. Although ViT attention maps can show improved alignment with ROIs when trained with MMPKD, their reliability as explanations remains opaque in both baseline and MMPKD models, as reflected by high variability and prior critiques \cite{chung2024evaluating,jain2019attention}. Attention map structures can vary widely—accurate and misleading maps coexist within and between methods and individual runs, highlighting the risk of overinterpreting qualitative attention visualizations. Future work should rigorously evaluate attention-based explainability and clarify conditions for effectively leveraging privileged modalities.

%% file: chapters/05_funding.tex
\section{Funding Statement}
This work was supported by the Senate of Berlin and the European Commision's Digital Europe Programme (DIGITAL) as grant TEF-Health (101100700).